\documentclass[sigconf, dvipsnames]{acmart}

\AtBeginDocument{%
  \providecommand\BibTeX{{%
    \normalfont B\kern-0.5em{\scshape i\kern-0.25em b}\kern-0.8em\TeX}}}

\setcopyright{acmcopyright}
\copyrightyear{2022} 
\acmYear{2022} 
\setcopyright{acmlicensed}\acmConference[MM '22]{Proceedings of the 30th ACM International Conference on Multimedia}{October 10--14, 2022}{Lisboa, Portugal}
\acmBooktitle{Proceedings of the 30th ACM International Conference on Multimedia (MM '22), October 10--14, 2022, Lisboa, Portugal}
\acmPrice{15.00}
\acmDOI{10.1145/3503161.3547747}
\acmISBN{978-1-4503-9203-7/22/10}

\acmSubmissionID{11}

\usepackage[T1]{fontenc}
\usepackage{fontawesome} 
\usepackage{multirow}
\usepackage{caption}
\usepackage{subcaption}

\begin{document}

\title[DrawMon]{DrawMon: A Distributed System for Detection of Atypical Sketch Content in Concurrent Pictionary Games}

\author{Nikhil Bansal}
\affiliation{%
  \institution{Centre for Visual Information Technology\\International Institute of Information Technology Hyderabad}
  \city{Hyderabad}
  \country{INDIA}
}
\email{nikhil.bansal@research.iiit.ac.in}

\author{Kartik Gupta}
\affiliation{%
  \institution{Centre for Visual Information Technology\\International Institute of Information Technology Hyderabad}
  \city{Hyderabad}
  \country{INDIA}
}
\email{kartik.gupta0204@gmail.com}

\author{Kiruthika Kannan}
\affiliation{%
  \institution{Centre for Visual Information Technology\\International Institute of Information Technology Hyderabad}
  \city{Hyderabad}
  \country{INDIA}
}
\email{kiruthika.k@research.iiit.ac.in}

\author{Sivani Pentapati}
\affiliation{%
  \institution{Centre for Visual Information Technology\\International Institute of Information Technology Hyderabad}
  \city{Hyderabad}
  \country{INDIA}
}
\email{pentapatisivani27@gmail.com}

\author{Ravi Kiran Sarvadevabhatla}
\affiliation{%
  \institution{Centre for Visual Information Technology\\International Institute of Information Technology Hyderabad}
  \city{Hyderabad}
  \country{INDIA}
}
\email{ravi.kiran@iiit.ac.in}

\renewcommand{\shortauthors}{Nikhil Bansal et al.}

\begin{abstract}
Pictionary, the popular sketch-based guessing game, provides an opportunity to analyze shared goal cooperative game play in restricted communication settings. However, some players occasionally draw atypical sketch content. While such content is occasionally relevant in the game context, it sometimes represents a rule violation and impairs the game experience. To address such situations in a timely and scalable manner, we introduce \textsc{DrawMon}, a novel distributed framework for automatic detection of atypical sketch content in concurrently occurring Pictionary game sessions. We build specialized online interfaces to collect game session data and annotate atypical sketch content, resulting in \textsc{AtyPict}, the first ever atypical sketch content dataset. We use \textsc{AtyPict} to train \textsc{CanvasNet}, a deep neural atypical content detection network. We utilize \textsc{CanvasNet} as a core component of \textsc{DrawMon}. Our analysis of post deployment game session data indicates \textsc{DrawMon}'s effectiveness for scalable monitoring and atypical sketch content detection. Beyond Pictionary, our contributions also serve as a design guide for customized atypical content response systems involving shared and interactive whiteboards. Code and datasets are available at \url{https://drawm0n.github.io}.
\end{abstract}



\begin{teaserfigure}
\centering
  \includegraphics[width=\textwidth]{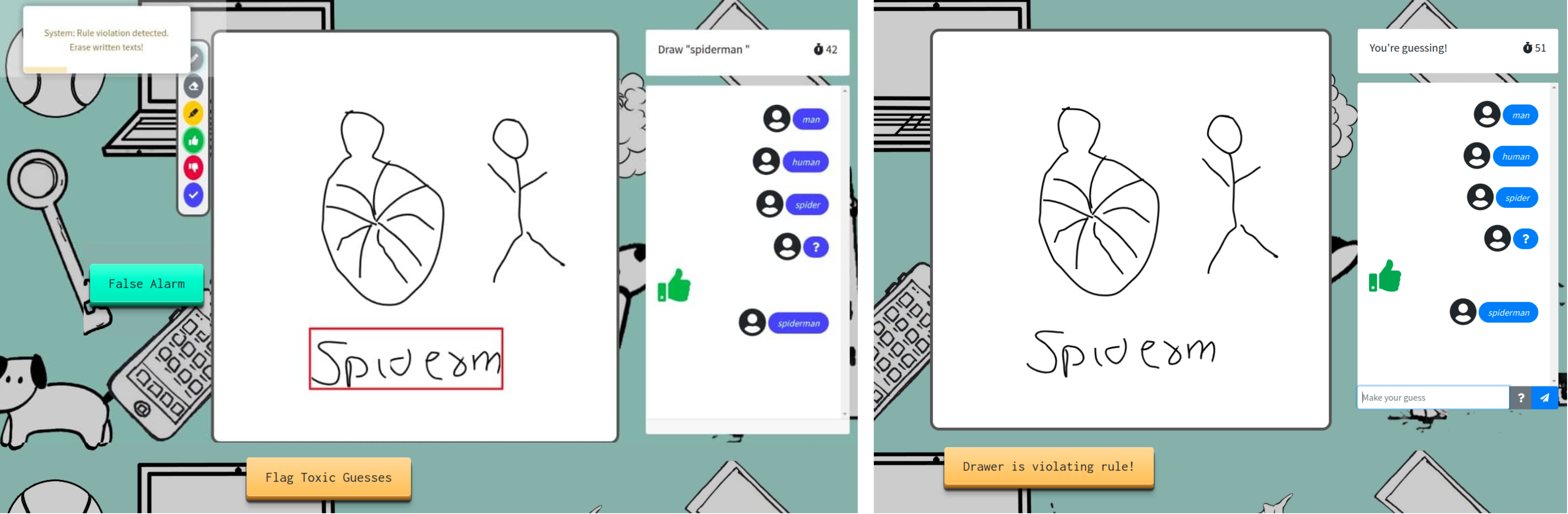}
  \caption{Screenshots of our game portal showing Drawer (left) and Guesser (right) activity during a Pictionary game. In this case, the Drawer has violated the game rules by writing text (`Spiderm') on the canvas. An automatic alert notifying the player (see top left of screenshot) and identifying the text location (red box on canvas) is generated  by our system \textsc{DrawMon}.}
    \label{fig:pictguess}
\end{teaserfigure}

\maketitle

\section{Introduction}

Shared digital whiteboards are becoming increasingly popular in educational and workplace settings as a natural mechanism for collaboration and communication~\cite{jensen2018remediating,fang2019augmented,davidson2021drawn,perlich2018cooperative,10.1145/3059454.3059477,10.1145/1054972.1055046}. The sharing aspect offers tremendous scope for interaction and a richer session experience. Unfortunately, shared whiteboards sometimes present situations where malicious participants draw controversial content~\cite{gameemblem}. Such activities tend to impair the collective experience of participants. Therefore, it is important to have scalable mechanisms for efficiently identifying and responding to such activities.

The popular social sketching game of Pictionary\textsuperscript{\texttrademark}, which we employ as a use case in this paper, also presents scenarios involving atypical sketched content. Pictionary is a wonderful example of cooperative game play to achieve a shared goal in communication-restricted settings~\cite{10.1145/3313831.3376316,10.1145/1378704.1378719,10.1145/3173574.3173767}. The game consists of a time-limited episode involving two players - a Drawer and a Guesser. The Drawer is tasked with conveying a given target phrase to a counterpart Guesser by sketching on a whiteboard~\cite{fay2017deconstructing}. The larger the number of target phrases correctly identified and the earlier the phrases are identified from the drawn sketch, greater the number of points accrued for the participating players. The rules of Pictionary forbid the Drawer from writing text on the whiteboard. This is usually not an issue when players are physically co-located. In the anonymized, web-based version of the game, however, the Drawer may cheat by writing text related to the target word on the digitally shared whiteboard, thus violating the rules. Intervention is possible by physically monitoring game sessions. However, such manual intervention is impractical and not scalable to an online setting involving a large number of multiple concurrent game sessions. Providing user interface options for player-triggered flagging of rule violation is another possibility. But such mechanisms are not completely reliable since the Guesser benefits from the content written on the canvas and does not have real incentive to use the flagging mechanism. 

\begin{figure*}[t!]
  \centering
  \includegraphics[width=\textwidth]{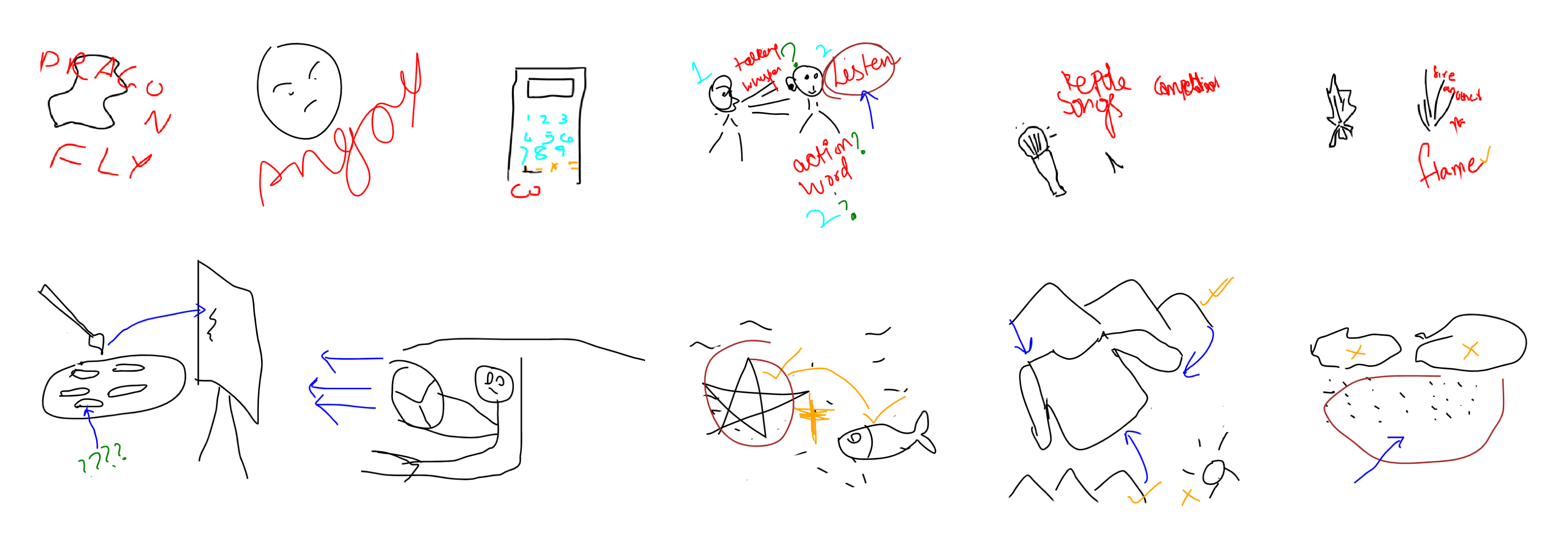}
  \caption{Some examples of atypical sketch content in Pictionary game sessions are shown as canvas screenshots. The content instances span \textcolor{red}{text}, \textcolor{cyan}{numbers}, \textcolor{ForestGreen}{question marks}, \textcolor{blue}{arrows}, \textcolor{Maroon}{circles} and other \textcolor{orange}{icons} (e.g. tick marks, addition symbol) categories - refer to Sec~\ref{sec:atypical-data} for details.}
  \label{fig:ano-ex}
\end{figure*}

Apart from malicious game play, atypical sketch content can also exist in non-malicious, benign scenarios. For instance, the Drawer may choose to draw arrows and other such icons to attract the Guesser's attention and provide indirect hints regarding the target word (see Fig.~\ref{fig:ano-ex}). Accurately localizing such activities can aid statistical learning approaches which associate sketch-based representations with corresponding target words~\cite{8509167}. Considering both malicious and benign scenarios, the broad requirement is for a framework which can respond to a variety of atypical whiteboard sketch content in a reliable, comprehensive and timely manner. To this end, we make the following contributions:

\begin{itemize}
    \item We introduce \textsc{AtyPict} - the first ever dataset of atypical whiteboard content.
    \item We introduce \textsc{DrawMon}, a distributed system for sketch content-based alert generation (Sec.~\ref{sec:drawmon}). We analyze sessions with \textsc{DrawMon} deployed for  Pictionary setting and demonstrate its effectiveness (Sec.~\ref{sec:drawmonuserstudy}).
\end{itemize}

Although presented in a Pictionary game context, our contributions serve as a design guide for developing  response frameworks involving shared and interactive whiteboards. For code, models and additional details, visit the project page \url{https://drawm0n.github.io}

\section{Related Work}

\noindent \textbf{Detecting and Flagging Anomalous Gameplay:} 
Some approaches employ a diverse mix of techniques for detecting cheating in online games~\cite{10.1007/978-3-642-38577-3_53,10.1145/1852611.1852643}. Dinh et al.~\cite{7725645} use hand-crafted game features and unsupervised machine learning approaches for offline detection of anomalous behavior. In our work, we introduce automatic deep learning based detection and flagging of anomalous gameplay in Pictionary. However, our system is also designed to detect secondary non-anomalous canvas entities which can potentially aid statistical understanding of canvas contents. 

\noindent \textbf{Sketch datasets:} Existing sketch datasets (e.g. TU-Berlin~\cite{eitz2012hdhso}, Sketchy~\cite{sketchy2016}, QuickDraw~\cite{jongejan2016quick}) have been created primarily in the context of sketch \textit{object} recognition problem -- assign a categorical label to a hand-drawn sketch. The category labels correspond to objects (nouns). Therefore, these datasets lack abstract sketches which tend to be drawn when words from other parts of speech (verbs, adjectives) are provided as targets. Existing datasets are also unnatural because they do not include canvas actions such as erase strokes or location emphasis. Also, no intermediate guess words are associated with sketched content. For a similar reason, these datasets do not contain atypical activities unlike the dataset we introduce. Sarvadevabhatla et al.~\cite{8509167} explore neural network based generation of human-like guesses, but for pre-drawn object sketches. However, they do not accommodate interactivity and non-sketch drawing canvas activities (e.g. erase, pointing emphasis). The Kondate dataset~\cite{6981047} contains on-line handwritten patterns of text, figures, tables, maps, diagrams etc. The OHFCD dataset~\cite{Awal2011FirstEO} pertains to online handwritten flowcharts.  Although challenging in their own way, these datasets are considerably more structured than our setting. Additionally, they share the sketch datasets' shortcoming of being too cleanly curated because actions such as erase are absent. As a unique aspect, our combination of a game setting and a time limit unleashes greater diversity and creativity, causing sketches in our dataset to be more spontaneous and less homogeneous compared to existing datasets.

\noindent \textbf{Detecting canvas items:} Recognizing atypical activities can be thought of as a stroke segmentation problem wherein each sketch stroke is labelled as either belonging to an atypical class or the default class (drawing). Stroke segmentation has been employed for labelling parts in object sketches either from stroke sequence information~\cite{yang2020sketchgcn,8784880,kaiyrbekov2019strokebased,qi2019sketchsegnet+} or within an image canvas~\cite{wang2020efficient,li2018fast}. 

Recognizing atypical sketch content can also be posed as an object detection problem. In this case, the objective is to obtain 2-D spatial bounding boxes enclosing sketch strokes corresponding to the atypical content. We adopt this approach because it is faster and more amenable to near real-time operation compared to segmentation. Handwritten text is the most common atypical sketch content class in Pictionary. Hence, it is reasonable to consider approaches solely designed for text detection in domains such as outdoor scenes and documents~\cite{jaderberg2014reading,Deng_2019,liu2019omnidirectional,Liu_2016,liao2018textboxes++,zhou2017east,Baek_2019_CVPR}. Similarly, detection-based approaches have been proposed for mixed graphic structures~\cite{8395187,schafer2019arrow,Elyan2020DeepLF}. However, graphic elements in these scenarios are  more structured compared to our Pictionary setting. 

\noindent \textbf{Pictionary-like guessing games:} Borrowing terminology from the seminal work of von Ahn and Dabbish~\cite{10.1145/1378704.1378719}, Pictionary can be considered an `inversion game' with full transparency. Riberio and Igarashi~\cite{10.1145/2380116.2380154} employ a sketching-based interactive guessing game to progressively learn visual models of objects. A review of Pictionary-like word guessing games involving drawing can be found in the work by Sarvadevabhatla et al.~\cite{8509167}.  In general, most of the existing works are confined to idealized toy settings~\cite{huang2020scones}, with some not even containing any sketching aspect~\cite{clark2021iconary,El-Nouby_2019_ICCV}. Unlike what we propose in this paper, they do not discuss the possibility of atypical content.
 
\section{Data Collection}
\label{sec:data-collection}

\subsection{Game Sessions}
\label{sec:game-browser}

Our browser-based game portal (Figures \ref{fig:pictguess},\ref{fig:pictguess2}) is compatible with mouse and touch inputs, scalable and can handle up to $50$ multiple concurrent Pictionary sessions. Consent is obtained and game instructions are provided when a player accesses the system for the first time. Players are assigned random names and paired randomly as Drawers and Guessers. The targets provided to the Drawers are sampled from a dictionary of $200$ guess phrases. We re-emphasize that the target phrases can be nouns (e.g. airplane, bee, chair), verbs (e.g. catch, call, hang) or adjectives (e.g. happy, lazy, scary). To ensure uniform coverage across the dictionary, the probability of a guess phrase being selected for a session is inversely proportional to the number of times it has been selected for elapsed sessions. The game has a time limit of $120$ seconds. The game ends when the Guesser enters a word deemed `correct' by the Drawer or when the time limit is reached. 

For the Guesser, a text box is provided for entering guess phrases. For the Drawer, the interface provides a canvas with tools to draw, erase and highlight locations (via a time-decaying spatial animation `ping') for emphasis (see Fig.~\ref{fig:pictguess}). In addition, \faThumbsOUp $\:$ and \faThumbsODown $\:$ buttons enable Drawer to provide `hot/cold' feedback on guesses. A question (\faQuestionCircleO) button is provided to the Guesser for conveying that the canvas contents are not informative and confusing. The canvas strokes are timestamped and stored in Scalable Vector Graphic (SVG) format for efficiency. In addition to canvas strokes (drawing and erasure related), secondary feedback activities mentioned previously (\faThumbsODown$\:$,\faThumbsOUp $\:$, \faQuestionCircleO, highlight) are also recorded with timestamps as part of the game session. 

Via our portal, we successfully gathered $3220$ timestamped episodes of diverse, realistic game play involving a total of $479$ participants in a large age range ($14$ years to $60$ years) and educational demographics (middle and high school students, graduate and undergraduate university students and working professionals). Please refer to project page for sample videos of game sessions, architectural overview of the game play system and plots with additional game session statistics.

\begin{table}[!t]
    \resizebox*{\linewidth}{!}
    {
    \begin{tabular}{llccc}
    \toprule
    \multicolumn{2}{c}{Sketch Content Type}& Number of & Number of & Number of \\
    \multicolumn{2}{c}{class} & occurrences & sessions & target phrases \\
    &&&containing&containing\\
    \hline
    \multicolumn{2}{l}{\textbf{\textit{Text}}}&\textbf{2419} & \textbf{478} & \textbf{180} \\
    \cline{3-5}
    &Individual letter&2244& 460& 178\\
    &Running hand&175& 103& 81\\
    \hline
    \multicolumn{2}{l}{\textbf{\textit{Numbers}}}&\textbf{331}& \textbf{73}& \textbf{28}\\
    \hline
    \multicolumn{2}{l}{\textbf{\textit{Circles}} }& \textbf{110}& \textbf{90}&\textbf{ 67}\\
    \hline
    \multicolumn{2}{l}{\textbf{\textit{Iconic}}} & \textbf{750} &\textbf{377} &\textbf{ 147}\\
    \cline{3-5}
    &Arrow & 497& 292& 129\\
    &Question mark & 158& 116& 78\\
    &Miscellaneous&95&54& 37\\
  \bottomrule
\end{tabular}
}
\caption{Statistics of atypical sketch content categories in game sessions.}
\label{tab:ano}
\end{table}

\subsection{Atypical Data}
\label{sec:atypical-data}
An atypical sketch content instance can be thought of as a subsequence of sketch curves relative to the larger sequence of curves that comprise the game session. We first describe the categories of atypical content usually encountered in Pictionary sessions: 
 \begin{itemize}
     \item \textit{Text}: Drawer directly writes the target word or hints related to the target word on the canvas. 
     \item \textit{Numerical}: Drawer writes numbers on canvas.
     \item \textit{Circles}: Drawers often circle a portion of the canvas to emphasize relevant or important content. 
     \item \textit{Iconic}: Other items used for emphasizing content and abstract compositional structures include drawing a question mark, arrow and other miscellaneous structures (e.g. double-headed arrow, tick marks, addition symbol, cross) and striking out the sketch (which usually implies negation of the sketched item).
 \end{itemize}
 
 Examples can be viewed in Fig.~\ref{fig:ano-ex}. It is important to remember that we consider only \textit{Text} writing as a rule violation in Pictionary. Other categories mentioned above are atypical but their presence is not considered a violation of game rules.

 \begin{figure}[!t]
   \includegraphics[width=\linewidth]{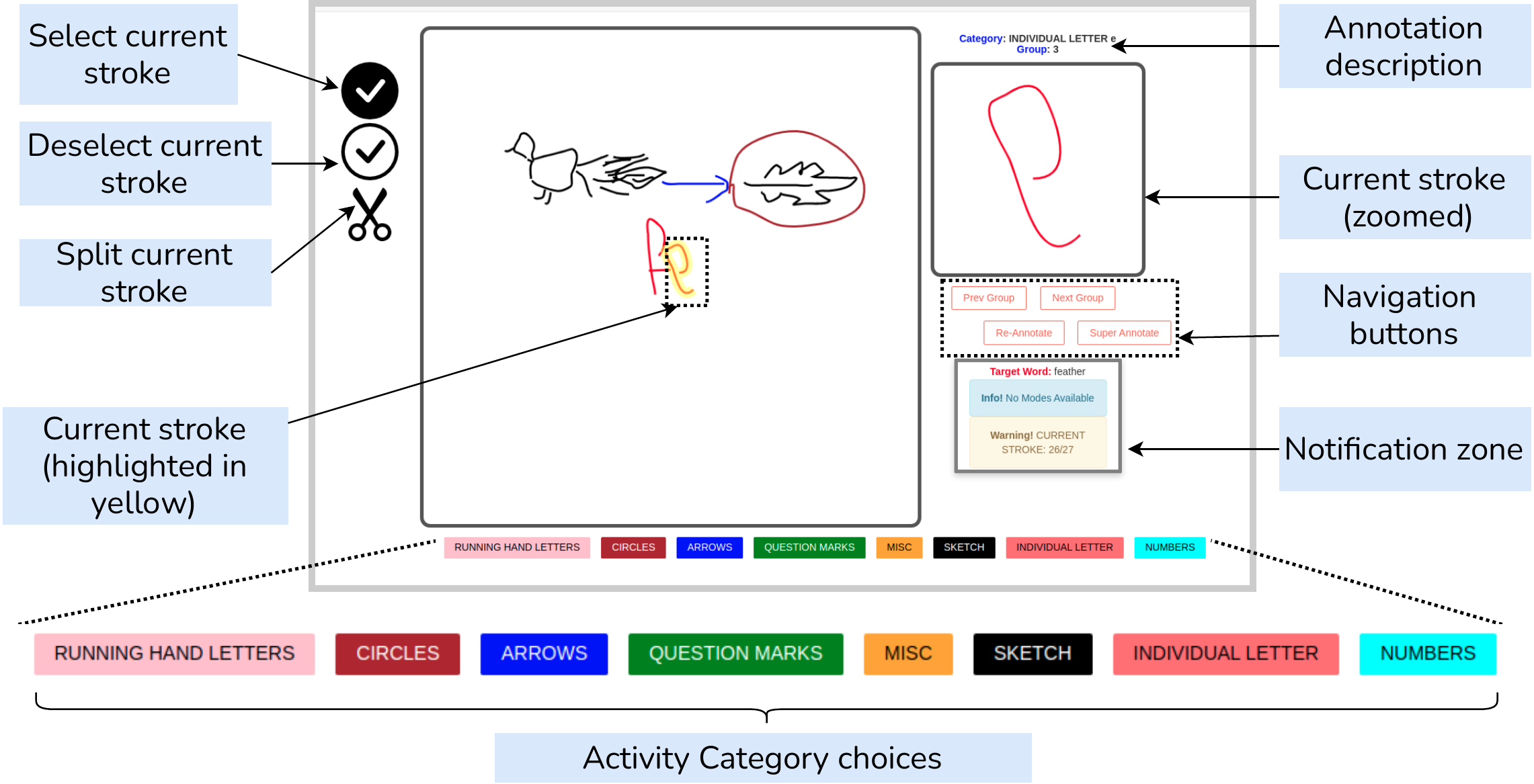}
   \caption{An illustration of annotation using the \textsc{CanvasDash} interface. 
   } 
   \label{fig:Dashboard}
 \end{figure}
To annotate atypical content, we use our custom-designed, browser-based annotation and visualization tool dubbed \textsc{CanvasDash} (see Fig.~\ref{fig:Dashboard})  - please refer to project page for details.

Using the described annotation procedure, we obtain our atypical Pictionary sketch dataset \textsc{AtyPict}. The occurrence statistics of atypical sketch categories across game sessions can be viewed in Table~\ref{tab:ano}. Representative visual examples can be viewed in Fig.~\ref{fig:ano-ex}.  Although we had earlier defined atypical content in terms of curve subsequences, the illustrations in Fig.~\ref{fig:ano-ex} show that the content can have a defined 2-D spatial extent and context relative to the canvas. For ease of processing, we consider this latter interpretation. In other words, we consider atypical content instances to be category-labelled 2-D spatial patterns sketched over time on the canvas. In terms of Fig.~\ref{fig:ano-ex}, detecting such content therefore corresponds to accurately detecting and categorizing 2-D spatial extents of entities shown color-coded on the canvas. 

\begin{figure*}[t!]
  \includegraphics[width=0.8\textwidth]{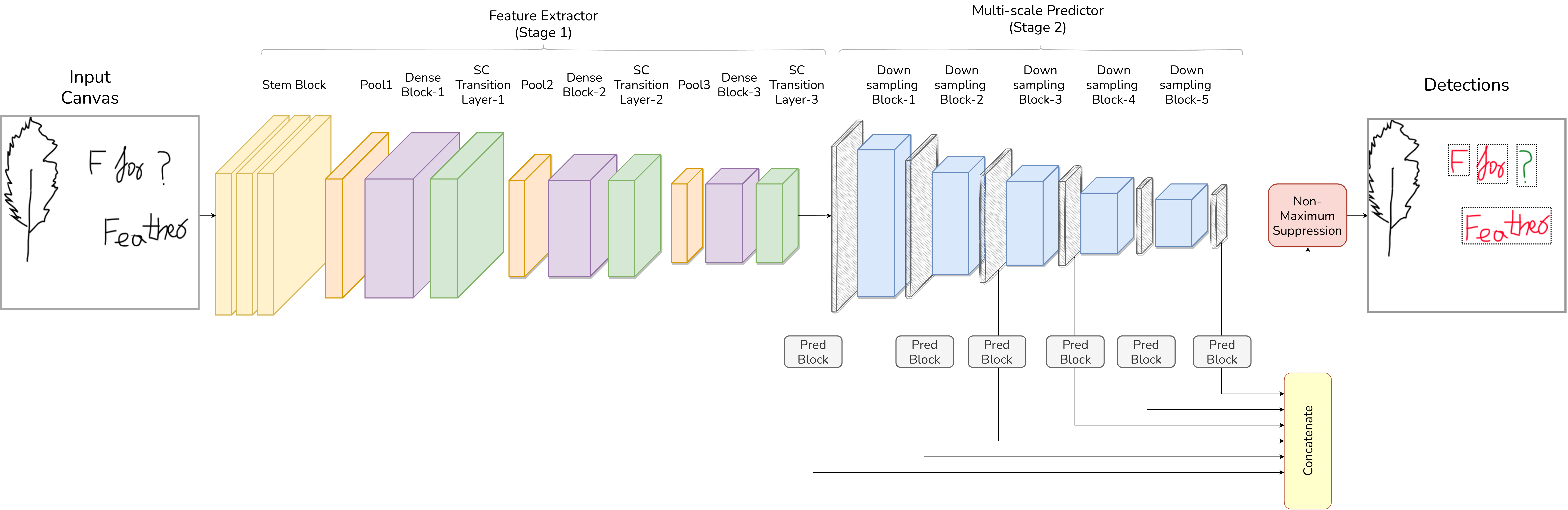}
  \caption{Architecture of \textsc{CanvasNet} deep neural network. Refer to Sec.~\ref{sec:canvasnetdeepnet} for details.}
  \label{fig:networkdiagram}
\end{figure*}

\section{CanvasNet}
\label{sec:canvasnet}

An effective approach for detecting atypical sketch instances needs to tackle the diversity in scale and appearance of various categories - a glance at Fig.~\ref{fig:ano-ex} makes this amply clear. In addition, the approach needs to utilize spatial context and be robust to the presence of similar looking yet semantically distinct regular (sketch) canvas content. To meet these requirements, we cast the problem as image-based object detection. In our case, the drawing canvas containing accumulated sketch strokes is the image. Any atypical content instances (e.g. \textit{Text}) present are considered spatially localized 2-D objects to be detected. For the object detection, we design a novel deep neural network which we dub \textsc{CanvasNet}. Before delving into details of \textsc{CanvasNet}, we first describe the data setup employed for its training and evaluation.

\subsection{Data Preparation}
\label{sec:strokedatapreparation}

As mentioned previously, the canvas elements for a given game session are represented as timestamped SVG curve elements. Each SVG element is either a  drawing stroke or an erasure stroke. We group drawing strokes into subsequences which are separated by erase strokes. The curves are converted to a 2-D point sequence representation and adaptively downsampled into line-based strokes using Ramer–Douglas–Peucker~\cite{RAMER1972244} algorithm  ($\epsilon=2$). The strokes are rendered on a $512 \times 512$ 2-D canvas with a stroke thickness of $4$ for the purpose of data annotation and representation. Note that a drawing stroke either belongs to one of the atypical classes (Sec.~\ref{sec:atypical-data}) or is a normal sketch stroke.  
The spatial extents of labelled stroke subsequences are used to automatically generate ground-truth data for training and evaluation of \textsc{CanvasNet} (Sec.~\ref{sec:canvasnet}). Examples of ground-truth bounding boxes for atypical categories can be seen as solid (non-dashed/non-dotted) rectangles in Fig.~\ref{fig:CanvasNetDetections}.

\noindent \textbf{Data Augmentation:} The number of game sessions containing atypical instances are considerably smaller compared to the total number of game sessions. To increase the amount of data available for deep network training in a realistic manner, we first isolate atypical instance stroke subsequences. We sample from this set and add the resulting subsequences to other sessions which share the same target phrase, but do not contain any atypical content. The atypical content subsequences are also randomly rotated and localized carefully. This ensures they are spatially disjoint from strokes of the reference game sessions (which originally lack such atypical entities) -- examples can be viewed in project page.

\subsection{\textsc{CanvasNet} Deep Network}
\label{sec:canvasnetdeepnet}

Inspired by the success of deep networks which attempt to detect text in photos~\cite{liu2016ssd,shen2017dsod,liao2018textboxes++}, we adopt a similar efficient approach for our \textsc{CanvasNet} deep network to detect atypical sketch instances on a drawing canvas. \textsc{CanvasNet} consists of two stages.

\textbf{Feature Extractor:} The first stage consists of a stem block containing three $3 \times 3$ unit separable convolution layers~\cite{chollet2017xception}. Our choice of seperable convolution layers is motivated by the reduction in number of parameters and operations involved. The first layer in stem block uses a stride of $2$ to downsample the input features. The stem block is then followed by a repeating three segment structure consisting of (i) a $2 \times 2$ max pooling layer (ii) a $6$ layered dense block~\cite{huang2017densely} with a growth rate of $48$. Each of the $6$ layers consists of a $1 \times 1$ separable convolution followed by a $3 \times 3$ separable convolution (iii) a $1 \times 1$ separable convolution layer denoted as SC Transition layer. This three segment structure is then repeated three times with similar parameters - see project page for additional details. 

\textbf{Multi-scale Predictor:} Atypical content (e.g. handwritten text, arrows) can occupy varying amounts of drawing canvas area, to detect them we use multi-scale predictor for prediction on multiple scales of feature maps~\cite{liao2018textboxes++, shen2017dsod}. For our setting, we use a customized multi-scale predictor for both handwritten text and non-text classes. The second stage sub-network is responsible for generating multi-scale feature maps and generating predictions over each of the feature maps. The output of third segment structure of the Feature Extractor is considered the first scale of the multi-scale feature map. The other feature map scales are obtained as outputs of successive downsampling blocks applied to the Feature Extractor's output (Fig.~\ref{fig:networkdiagram}) - see project page for additional details. The feature maps are passed individually through a prediction block. The multi-scale prediction features are concatenated and non-maximal suppression is applied to generate the final bounding box predictions.

\textit{Prediction block:} This consists of a $3 \times 5$ separable convolution layer, followed by a fully connected layer comprising the prediction. The rectangular filter dimensions ($3 \times 5$) used in the block ensure that elongated objects can be detected reliably. 

\textit{Anchor boxes with vertical offsets:} Among the atypical object categories, words have larger aspect ratios and range of box orientation. Therefore, we set anchor aspect ratios to $1, 2, 3, 4, 5, 1/2, 1/3, 1/5$ with $\pm0.25$ as the vertical offset.

\noindent \textbf{Optimization:} We formulate the loss function for \textsc{CanvasNet} as a combination of a classification loss $L_{cls}$ and a bounding-box localization loss $L_{loc}$:

\begin{equation}
    L = \alpha \left( \frac{1}{N} \sum_{i=1}^{N} L_{cls}(P_i, G_i) \right) + \left( \frac{1}{M} \sum_{i=1}^{N} \sum_{j=1}^{c} G_{ij}*L_{loc}(B_{i}, B^{gt}_{i}) \right)
\end{equation}

where $G$ is the ground truth label matrix ($G_{ij} = 1$ if $i$-th anchor box belongs to atypical category $j$, else $G_{ij} = 0$), $P$ is the predicted confidence score matrix ($P_{ij}$ indicates the confidence score that $i$-th anchor box belongs to category $j$), this means $P_i$ is a row of confidence scores for $i$-th box to belong to various atypical categories, similarly $G_i$ is a row where values for all atypical category is $0$ except for one(to which $i$-th box belongs).  
In the above loss function formulation, $B^{gt}_i$ and $B_i$ denotes the ground truth and predicted offsets for the $i$-th anchor box. $N$ is the total number of anchor boxes, $M$ is the total number of ground truth anchor boxes not belonging to background class, $c$ represents the number of atypical categories. We use focal loss\cite{lin2017focal} for $L_{cls}$ which prioritizes a sparse set of hard examples and prevents large number of negatives from overwhelming the detector. For localisation task, we adopt Distance-IoU Loss~\cite{zheng2020distance}:

\begin{equation*}
  \mathcal{L}_{D I o U}=1-I o U+\frac{\rho^{2}\left(\mathbf{b}, \mathbf{b}^{g t}\right)}{c^{2}}
\end{equation*}

where $b$ and $b^{gt}$ denote the central points of predicted and ground truth bounding box, $\rho^{2}\left(\mathbf{b}, \mathbf{b}^{g t}\right)$ gives the square of the Euclidean distance between them, and $c$ is the length of the diagonal of the smallest enclosing box covering the two bounding boxes.

\section{DrawMon}
\label{sec:drawmon}

\begin{figure}[t!]
  \centering
    \includegraphics[width=0.4\textwidth]{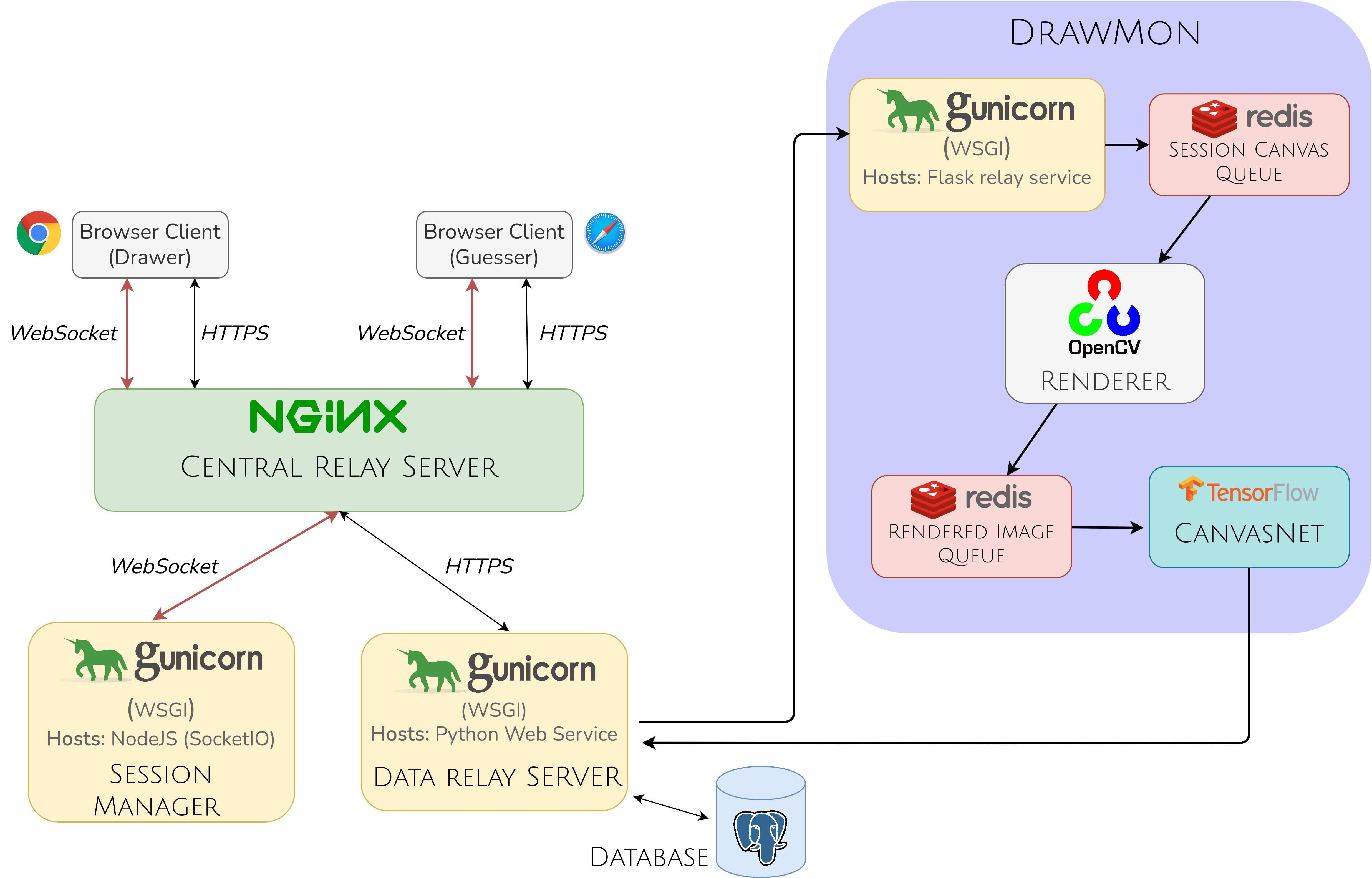}
    \caption{System architecture for Pictionary game setup and \textsc{DrawMon} (Sec.~\ref{sec:drawmon}).}
    \label{fig:pictguess2}
\end{figure}

\begin{figure*}[!t]
  \includegraphics[width=0.8\textwidth]{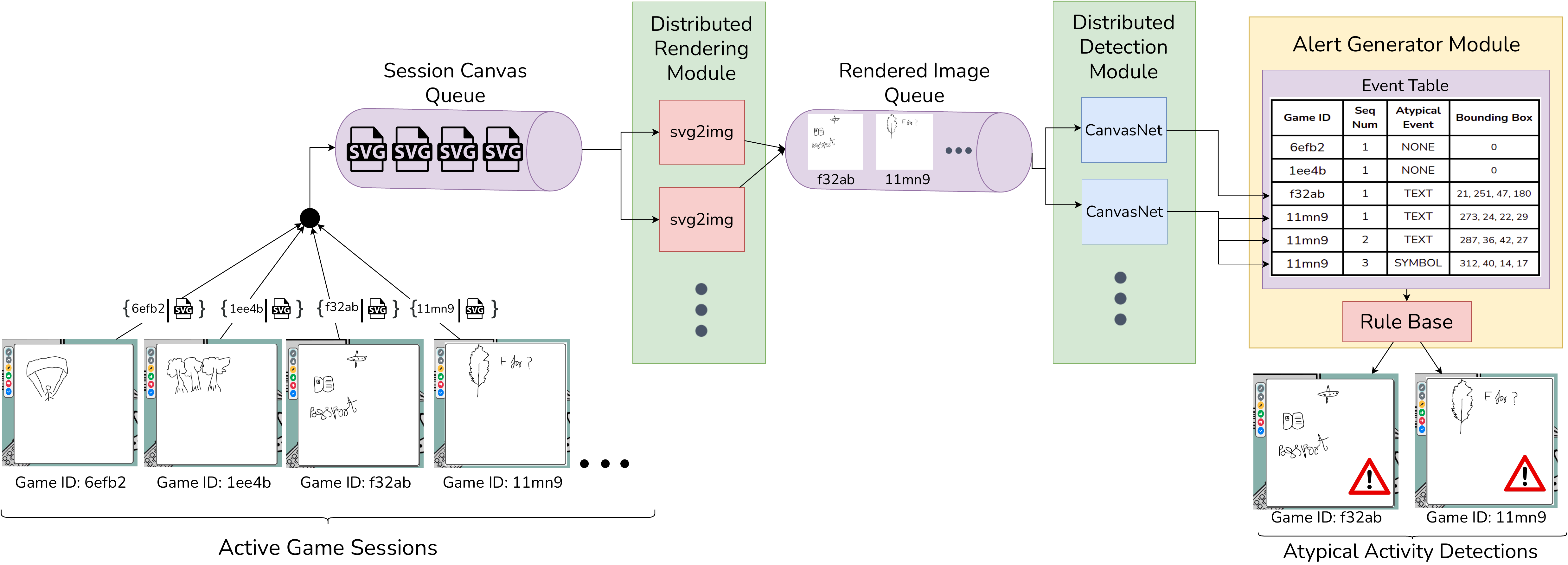}
  \caption{A pictorial overview of \textsc{DrawMon} - our distributed atypical sketch content response system (Sec.~\ref{sec:drawmon}). Also see Fig.~\ref{fig:pictguess2} for additional architectural details.}
  \label{fig:judge}
\end{figure*}

\begin{table*}[!t]
    \caption{Results for atypical content detection}
    \begin{subtable}[t]{.36\textwidth}
        \caption{\textsc{CanvasNet} performance for atypical content classes. IoU=0.5 refers to detection threshold. Refer to Sec.~\ref{sec:evalprotocol} for details.}
     \label{tab:multiclass}
        \raggedright
        \resizebox{\linewidth}{!}
{
           \begin{tabular}{c|cc}
\toprule
 \multirow{2}{*}{Atypical Content Category}& \multicolumn{2}{c|}{IoU=0.5} \\
 \cline{2-3}
      &  mAP & mAR \\
\hline
      Text & 0.58 & 0.80 \\
      Number & 0.44 & 0.61 \\
      Icon & 0.55 & 0.68 \\
      Circle & 0.72 & 0.85 \\
      \bottomrule
    \end{tabular}
    }
    \end{subtable}%
    \hfill
   \begin{subtable}[t]{.6\textwidth}
        \raggedleft
        \caption{Performance comparison with baselines. mAP = mean Average Precision, mAR = mean Average Recall, \#Parameters = the number of trainable weights in the corresponding deep network in millions, ADT = average detection time per image in milliseconds.}
  \label{tab:baselines}
        \resizebox{\linewidth}{!}
{
        \begin{tabular}{l|cc|cc|c|c}
    \toprule
    \multicolumn{1}{c|}{\multirow{2}{*}{Method}} & \multicolumn{2}{c|}{Text only} & \multicolumn{2}{c|}{Multiclass} & \# Parameters & ADT\\
    \cline{2-5}
    & mAP & mAR & mAP & mAR & M=million & (m.sec)\\
    \hline
    \textbf{\textsc{CanvasNet}} & \textbf{0.78} & \textbf{0.90 } & \textbf{0.41} & \textbf{0.53} & 1.90 M & 35\\
    \hline
    BiLSTM+CRF~\cite{darvishzadeh2019cnn} & 0.06 & 0.04 & 0.02 & 0.03 & 0.01 M &85\\
    
    SketchsegNet+\cite{qi2019sketchsegnet+} & 0.56 & 0.32 & 0.04 & 0.11 & 3.90 M &21\\
    
    Tiny-YOLOv4\cite{wang2020scaled} & 0.26 & 0.65 & 0.31 & 0.51 & 5.88 M & 40\\
    
    TextBoxes++\cite{liao2018textboxes++} & 0.41 & 0.65 & 0.25 & 0.39 & 29.31 M & 51\\
    
    DSOD\cite{shen2017dsod} & 0.43 & 0.66 & 0.25 & 0.40 & 17.49 M & 52\\
    
    CRAFT \cite{Baek_2019_CVPR} & 0.47 & 0.69 & 0.17 & 0.30 & 1.18 M & 34 \\
    
    \bottomrule
 \end{tabular}
 }
    \end{subtable}
\end{table*}

Consider a scenario with multiple online Pictionary game sessions in progress. We require a  framework for automatic and concurrent monitoring of these game sessions for any atypical activities (e.g. a rule violation such as writing text on canvas). Such a framework needs to be reliable, scalable and time-efficient. To meet these requirements, we propose \textsc{DrawMon} - a distributed alert generation system (see Fig.~\ref{fig:judge}). Each game session is managed by a central Session Manager which assigns a unique session id (Fig.~\ref{fig:pictguess2}). For a given session, whenever a sketch stroke is drawn, the accumulated canvas content (i.e. strokes rendered so far) is tagged with session id and relayed to a shared Session Canvas Queue. For efficiency, the canvas content is represented as a lightweight Scalable Vector Graphic (SVG) object. The contents of the Session Canvas Queue are dequeued and rendered into corresponding $512 \times 512$ binary images by Distributed Rendering Module  in a distributed and parallel fashion. The rendered binary images tagged with session id are placed in the Rendered Image Queue. The contents of Rendered Image Queue are dequeued and processed by Distributed Detection Module. Each Detection module consists of our custom-designed deep neural network \textsc{CanvasNet} which processes the rendered image as input. \textsc{CanvasNet} outputs a list of atypical activities (if any) along with associated meta-information (atypical content category, 2-D spatial location). 

The outputs from multiple distributed \textsc{CanvasNet} instances within the Distributed Detection Module are routed to the Alert Generator Module. An activity Record Table within this module records information related to ongoing game sessions and atypical content instances. This table is analyzed with respect to a Rule Base sub-module which generates appropriate alerts and relays them to the 
appropriate game sessions. Since rule violations are of predominant interest, other atypical content  alerts can be filtered out. Incoming alerts are finally displayed on the game session user interface (UI)  - see Fig.~\ref{fig:pictguess}.

Also note that two manually-controlled mechanisms related to alert generation exist within the game UI. The Guesser player can press a button labelled `Drawer is violating rule!'. This  simply generates the alert (but does not highlight the canvas location where violation occurs). On the Drawer's side, the button `False Alarm' can be used to dismiss false positive alerts (see Fig.~\ref{fig:pictguess}). In the current deployment, we utilize the \textit{Text} detection variant of \textsc{CanvasNet} to detect text writing event on canvas.

\section{Experiments and Results}
\label{sec:experiments}

We first describe the experiments and results for atypical content detection. Following standard machine learning protocols, we divide data into training, validation and test splits. For each target phrase, the sessions containing atypical content are randomly split in the ratio $70$ (train) : $15$ (validation) : $15$ (test). Since we perform data augmentation on atypical content-free sessions, we divide such sessions in the aforementioned ratio as well. The respective data splits are combined to obtain the final groups.

\subsection{Atypical content detection}
\label{sec:evalprotocol}

\noindent \textbf{Baselines:} All along, our approach for detecting atypical activities treats the canvas as a 2-D image. In effect, the game session is considered to be a video-like frame sequence of 2-D canvas images. We also consider alternate approaches wherein the game session is processed as a sequence of curves. Each curve is labelled either as a regular sketch stroke or one associated with an atypical content category. Briefly, the baselines we consider are the following:
\textit{BiLSTM+CRF~\cite{darvishzadeh2019cnn} - } This classifies each stroke in a sketch sequence as one of the atypical classes using a bidirectional Long Short Term Memory (BiLSTM) neural network~\cite{hochreiter1997lstm} and Conditional Random Field (CRF)~\cite{sutton2006introduction}. The input to this model is a set of hand-crafted features of the strokes. \textit{SketchSegNet+~\cite{qi2019sketchsegnet+} - } This classifies each point in a sketch sequence as one of the atypical classes using bidirectional LSTM. For image based models, we train appropriately modified versions of two state-of-the-art generic object detectors -- \textit{DSOD~\cite{shen2017dsod}} and \textit{Tiny-YOLOv4}~\cite{wang2020scaled}. We also train modified versions of two popular scene text detection models -- \textit{TextBoxes++~\cite{liao2018textboxes++}} and \textit{CRAFT~\cite{Baek_2019_CVPR}}. Please see project page for architectural details of baselines.

\noindent \textbf{Evaluation Protocol:} We conduct evaluation using two protocols. In the first protocol, we consider models trained to detect all atypical content classes. To score performance, we use the standard object detection measures --  mean-average-precision (mAP) and  mean-average-recall (mAR)~\cite{zhou2019iou}. These measures are typically reported on a $[0,1]$ scale -- larger the better. mAP and mAR are reported at an Intersection-over-Union (IoU) threshold of $0.5$. In other words, an overlap of 50\% or greater between predicted bounding box and ground-truth bounding box is deemed correct (assuming predicted category label also matches). 

\noindent \textbf{Training:} For training \textsc{CanvasNet}, we employ Adam optimizer~\cite{kingma2014adam} with a mini-batch size of $8$, the exponential decay rate for $1st$ and $2nd$ moment estimates set to $0.9$ and $0.999$ respectively. We stop training after $50$ epochs. Classification loss weight ($\alpha$) is set to $1000$ for quick convergence. The training takes approximately $4.5$ hours on two GTX 1080Ti 11GB GPUs. We use \textit{mish}~\cite{misra2020mish} activation function since it provides fast convergence and improved performance compared to the standard \textit{relu} activation. For training the text-only model, we used \textit{hard mining training regime} wherein false-positive samples having IoU overlap value with ground truth in the range $[0.45, 0.55]$ are chosen for mining. We observed significant increase in $mAP$, $mAR$ compared to regular training regime due to this regime. For training the multiclass model, we used mini-batch re-sampling with class balancing to counteract the presence of imbalanced per-class sample distribution.

\begin{figure*}[!t]
    \centering
    \includegraphics[width=0.9\textwidth]{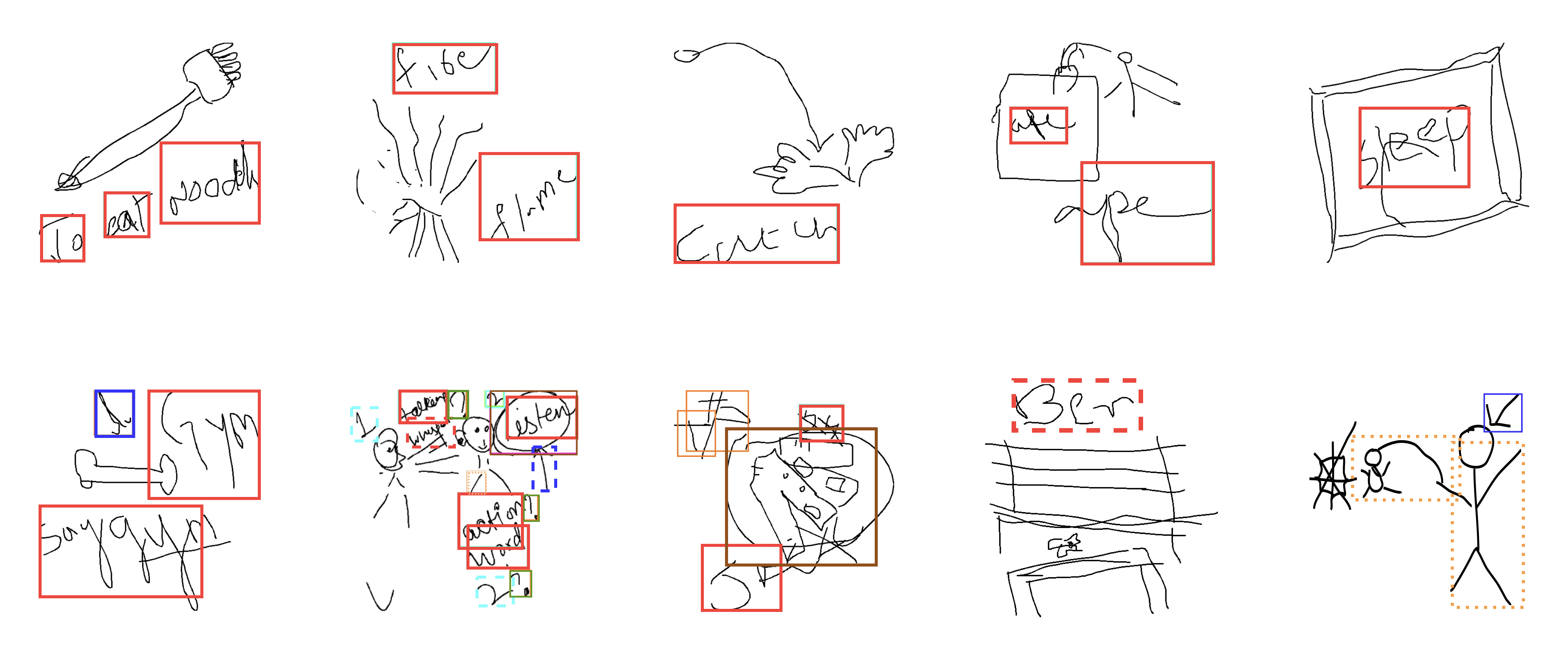}
    \caption{Examples of atypical content detection by \textsc{CanvasNet}. False negatives are shown as dashed rectangles and false positives as dotted rectangles. Color codes are: \textcolor{red}{text}, \textcolor{cyan}{numbers}, \textcolor{ForestGreen}{question marks}, \textcolor{blue}{arrows}, \textcolor{Maroon}{circles} and other \textcolor{orange}{icons} (e.g. tick marks, addition symbol). }
    \label{fig:CanvasNetDetections}
\end{figure*}

\noindent \textbf{Detection Results:} \textsc{CanvasNet}'s performance for all the atypical categories can be viewed in Table~\ref{tab:multiclass}. Fig.~\ref{fig:CanvasNetDetections} depicts examples of \textsc{CanvasNet} detections, including some failure cases. We conducted ablation experiments to determine the relative importance of our architectural and optimization choices. Details of the ablation configurations and results can be viewed in the project page. The comparative evaluation of \textsc{CanvasNet} with baseline approaches is summarized in  Table~\ref{tab:baselines}. Note that the comparison also includes compute-related aspects (number of trainable parameters in the approaches and average detection time).

Table~\ref{tab:multiclass} shows \textsc{CanvasNet}'s performance for various atypical content categories. The consistent depictions of \textit{Circle} enables good performance for the category. Detecting isolated numbers is slightly more challenging. Empirically, we observed that sketched content resembling letters (e.g. a mountain sketch) or numerals (e.g. vertical bar groups) accounted for most of the false positive detections. Text spanning a significantly large extent of the canvas and unusually oriented numbers accounted for majority of the missed detections (false negatives). Performance scores for ablative variants of \textsc{CanvasNet} are included in project page.

From Table~\ref{tab:baselines}, we see that \textsc{CanvasNet} clearly outperforms a variety of   baseline approaches (Sec.~\ref{sec:evalprotocol}). This is predominantly due to the carefully considered architectural and optimization choices in designing \textsc{CanvasNet}. The results also illustrate the superiority of image-based approaches compared to the sketch stroke processing approaches (\textit{BiLSTM
+CRF, SketchSegNet+}). Keeping the rule-violation detection scenario in mind, we also trained variants designed to detect the single class \textit{Text}. As the `Text only' column in Table~\ref{tab:baselines} shows, \textsc{CanvasNet} remains the best performer. Consequently, we utilize this model variant as part of \textsc{DrawMon} in our game deployment scenario (Sec.~\ref{sec:drawmon}). From the table (column named `Parameters'), we also note that \textsc{CanvasNet} achieves its superior performance despite containing a smaller number of parameters relative to most of the baselines. 

\subsection{DrawMon User Study Experiments}
\label{sec:drawmonuserstudy}

To quantify the efficacy of \textsc{DrawMon}, we analyzed game session data with \textsc{DrawMon} deployed to detect text. The canvas contents are relayed to \textsc{DrawMon} every $1$ second. We deployed $4$ \textsc{CanvasNet} instances within the Distributed Detection Module on two 2080Ti GPUs alongside $16$ worker processes for svg to image conversion. The combined peak usage of GPU RAM was $20$ GB while peak CPU RAM usage was $15$ GB. $23$ participants ($11$ male, $12$ female) in the age group $19-25$ (mean=$20.8$, std.=$3.1$), recruited using social media and from the institution's student pool, participated in the study. Each session had an average duration of $47.5$ sec (std.= $37.4$) with the maximum being $120$ seconds. Over the study period, the maximum number of concurrent game sessions managed by \textsc{DrawMon} was $4$. From the resulting set of $145$ game sessions, $69$ contained atypical text activities. During the sessions, we recorded timestamped alerts from \textsc{DrawMon}, false alarm notifications by the Drawer player and rule violation notifications from the Guesser player. The results from the study are summarized in Table~\ref{tab:userstudy}. To determine \textsc{DrawMon}'s throughput, we measured two quantities. The first, processing time (\texttt{p-time}), is the average elapsed time between the canvas representation being sent to \textsc{DrawMon} and receiving an alert. In case no alerts were generated, the timestamp corresponding to end of \textsc{CanvasNet} processing was considered. From our data, \texttt{p-time} was $0.4$s. The other measurement was the maximum number of concurrently active sessions (\texttt{n-sess}) -- this was $4$. Defining the effective throughput rate to be \texttt{tpr} = \texttt{p-time}/\texttt{n-sess}, we obtain an average processing rate of $10$ items per second.

\begin{table}[!t]
\captionof{table}{User study statistics with \textsc{DrawMon} deployed.}
\label{tab:userstudy}
\resizebox{\linewidth}{!}
{
\centering
    \begin{tabular}{lc}
    \toprule
    Game Event Type & Count \\
    \midrule
    (True Positive) \textsc{DrawMon} generates `Rule Violation' alert. Drawer doesn't press `False Alarm' button. & $62$ \\
    (False Positive) \textsc{DrawMon} generates `Rule Violation' alert. Drawer presses `False Alarm' button.   & $32$ \\ 
    (False Negative) No `Rule Violation' alert. Guesser presses `Drawer is violating rule' button. & $6$ \\
    \bottomrule
    \end{tabular}
}
\end{table}

 \noindent \textbf{Results and Analysis:} The results from \textsc{Drawmon} deployment user study are summarized in Table~\ref{tab:userstudy}. From the table, we see that a significant fraction of \textsc{DrawMon} generated alerts are valid (see `True Positives'). From the results, \textsc{DrawMon}'s precision is $0.66$ while recall is $0.91$. Post the user study, we conducted a brief survey with Likert-type questions on a $1$ to $5$ scale with $5$ being the best. `Q: How responsive was \textsc{DrawMon} to valid rule violations?' : The average score was $3.63$ (s.d.=$0.74$), indicating reasonably high system throughput despite multiple concurrent sessions. This is also supported by the recorded throughput rate (\texttt{tpr}) mentioned previously in this section. `Q: How was the overall game experience?' : The score was $3.91$ (s.d.=$0.60$), suggesting a positive session experience and  satisfaction with rule violation detection and response mechanisms.
 
User study plots and sample videos of game sessions with \textsc{DrawMon} in action can be viewed in project page.

\subsection{Application Scenarios}
\label{sec:limitationsotherscenarios}

\noindent \textbf{Application Scenarios:} Although we have used Pictionary as a use case scenario, we expect \textit{DrawMon} to be suitable for other shared and interactive whiteboard scenarios. For instance, in a writing related setting, the notion of atypical categories can be the exact opposite of Pictionary scenario: text on canvas would be routine while drawings might be considered abnormal. This can be tackled by appropriate data labelling, for e.g. using our \textsc{CanvasDash} annotation tool, and subsequently retraining \textsc{CanvasNet} deep network. In another scenario, consider participants grouped into teams for a collaborative scene drawing task~\cite{davidson2021drawn,10.1145/3313831.3376517}. \textsc{DrawMon}, using a \textsc{CanvasNet} configured for sketched scene recognition~\cite{zou2018sketchyscene}, can alert the instructor on progress and task completion. For this task, a \textsc{CanvasNet} instance trained to recognize individual objects and iconic components from our dataset (e.g. arrows, `addition mark') could also be included as additional detection component for expanding the detection capability.

\section{Conclusion and Future Work}

\textsc{DrawMon} is a distributed framework for monitoring multiple shared interactive whiteboards for detecting atypical content.  We use a Pictionary-like sketching game as the use case scenario.  \textsc{DrawMon} is enabled by a number of equally important lateral contributions - (i) \textsc{CanvasDash} - an intuitive dashboard UI for annotation and visualization (ii) \textsc{AtyPict} - a first of its kind dataset for atypical sketch content (iii) \textsc{CanvasNet} - a deep neural network for atypical content detection. Together, these reusable contributions create the possibility of developing similar frameworks for other shared and interactive whiteboard scenarios. Apart from atypical content detection, we expect our game session dataset to be a valuable resource in itself for analyzing player characteristics and strategies in communication restricted non-adversarial games~\cite{10.1145/3313831.3376316}. In addition, we plan to develop practical AI agents which can mimic human Pictionary players in a more interactive, realistic manner compared to existing non-interactive works~\cite{bhunia2020sketch,huang2020scones}. 

\bibliographystyle{ACM-Reference-Format}
\bibliography{bibfile}

\end{document}